\documentclass[journal,twoside,web]{ieeecolor}
\usepackage{tmi}
\usepackage{cite}
\usepackage{amsmath,amssymb,amsfonts}
\usepackage{algorithmic}
\usepackage{graphicx}
\usepackage{textcomp}
\usepackage{multirow}

\usepackage{subcaption}
\usepackage{booktabs}
\usepackage{hyperref}
\usepackage{gensymb}
\usepackage{array}

\usepackage[nolist,nohyperlinks]{acronym}

\acrodef{us}[US]{ultrasound}
\acrodef{cnn}[CNN]{convolutional neural network}
\acrodef{mse}[MSE]{mean square error}
\acrodef{mri}[MRI]{magnetic resonance imaging}
\acrodef{ga}[GA]{gestational age}
\acrodef{dl}[DL]{deep learning}
\acrodef{ml}[ML]{machine learning}
\acrodef{dof}[DoF]{degree of freedom}
\acrodef{6d}[6D]{six-dimensional}
\acrodef{4d}[4D]{four-dimensional}
\acrodef{3d}[3D]{three-dimensional}
\acrodef{2d}[2D]{two-dimensional}
\acrodef{sp}[SP]{standard plane}
\acrodef{nhs}[NHS]{National Health System}
\acrodef{ai}[AI]{artificial intelligence}
\acrodef{csp}[CSP]{cavum septum pellucidum}
\acrodef{lv}[LV]{lateral ventricle}
\acrodef{tt}[TT]{transthalamic}
\acrodef{tv}[TV]{transventricular}
\acrodef{tc}[TC]{transcerebellar}
\acrodef{rnn}[RNN]{recurrent neural network}
\acrodef{sl}[SL]{supervised learning}
\acrodef{hc}[HC]{head circumference}
\acrodef{bmi}[BMI]{body mass index}
\acrodef{lfd}[LfD]{Learning from Demonstration}
\acrodef{loocv}[LOOCV]{Leave One Out Cross-Validation}
\acrodef{dicom}[DICOM]{Digital Imaging and Communications in Medicine}
\acrodef{roi}[ROI]{Region of Interest}
\acrodef{rms}[RMS]{Root Mean Square}
\acrodef{rl}[RL]{reinforcement learning}
\acrodef{iou}[IoU]{Intersection over Union}
\acrodef{usu}[USU]{Ultrasound Screening Unit}
\acrodef{ss-s}[SS-Seg]{semi-supervised segmentation}
\acrodef{s-s}[S-Seg]{supervised segmentation}
\acrodef{aspp}[ASPP]{Atrous Spatial Pyramid Pooling}
\acrodef{wBCE}[wBCE]{weighted binary cross-entropy}
\acrodef{BCE}[BCE]{binary cross-entropy}

\def\BibTeX{{\rm B\kern-.05em{\sc i\kern-.025em b}\kern-.08em
    T\kern-.1667em\lower.7ex\hbox{E}\kern-.125emX}}
\begin{document}
\title{Measuring proximity to standard planes during fetal brain ultrasound scanning}
\author{Chiara Di Vece, \IEEEmembership{Student Member, IEEE}, Antonio Cirigliano, Maela Le Lous, Raffaele Napolitano, Anna L. David, Donald Peebles, Pierre Jannin, Francisco Vasconcelos, and Danail Stoyanov, \IEEEmembership{Senior Member, IEEE}
\thanks{Manuscript received day month 2026. This work was supported in whole, or in part, by the Wellcome/EPSRC Centre for Interventional and Surgical Sciences (WEISS) [203145/Z/16/Z], the Department of Science, Innovation and Technology (DSIT), and the Royal Academy of Engineering under the Chair in Emerging Technologies programme.}
\thanks{C. Di Vece, A. Cirigliano, M. Le Lous, F. Vasconcelos, and D. Stoyanov are with the UCL Hawkes Institute and the Department of Computer Science, University College London (UCL), London WC1E 6AE, UK}
\thanks{M. Le Lous and P. Jannin are with the University of Rennes, INSERM, LTSI - UMR 1099, F35000, Rennes, France}
\thanks{R. Napolitano is with the Fetal Medicine Unit, Elizabeth Garrett Anderson Wing, UCLH NHS Foundation Trust, and the Elizabeth Garrett Anderson Institute for Women's Health, UCL, London, UK}
\thanks{A. L. David and D. Peebles are with the Department of Obstetrics and Gynecology, UCLH, London, UK}
\thanks{Corresponding author: {\tt\small chiara.divece.20@ucl.ac.uk}}}

\maketitle

\begin{abstract}
This paper presents a pipeline designed to bring ultrasound (US) plane pose estimation closer to clinical use, demonstrating the feasibility of continuous, real-time proximity feedback for navigation to the standard planes (SPs) in the fetal brain. We propose a semi-supervised segmentation model that uses labeled SPs and unlabeled slices from 3D US volumes (non-SPs), achieving 0.93 mean Intersection over Union (mIoU) on SPs and 0.86 mIoU on arbitrary non-SPs. The model incorporates a classification mechanism to identify and filter out frames lacking the fetal brain, and to generate masks for those containing it, enhancing the relevance of plane pose regression in clinical settings. Combined with 6D plane pose regression, our pipeline provides sensorless, continuous proximity detection to SPs with real-time distance metrics rather than binary plane recognition. Furthermore, we validate its translational viability by deploying the system on an NVIDIA Clara AGX edge device, achieving a real-time inference speed of 39 Hz, which exceeds standard clinical acquisition rates. Unlike prior methods validated on curated volume slices, we evaluate the pipeline retrospectively on real fetal scan videos from 17 sonographers of varying expertise: operators freeze near, rather than exactly at, the local minima of the proximity signal, consistent with clinical freeze-timing behavior, whereas proximity alone does not predict expert SP quality scores. The approach complements existing fetal US technologies and is a step toward image-based navigation support in prenatal scanning.
\end{abstract}

\begin{IEEEkeywords}
Ultrasound, Fetal brain, Semi-supervised learning, Plane pose estimation 
\end{IEEEkeywords}

\section{Introduction}
\label{sec:introduction}

\IEEEPARstart{F}{etal} \ac{us} is instrumental in detecting fetal growth abnormalities, potentially averting adverse outcomes for both mother and child~\cite{Hadlock1985}. Accurate assessment relies on the sonographer's ability to acquire and interpret standard imaging planes, known as \acp{sp}. Navigating these planes requires spatial awareness, precise hand-eye coordination, and anatomical knowledge. This task is often challenging for novice sonographers, who struggle to manipulate the transducer in three dimensions while interpreting the resulting \ac{2d} images as representations of \ac{3d} structures~\cite{Sarris2011,Bahner2016}. Consequently, operator experience variability causes diagnostic discrepancies, with novices at higher risk of missing key diagnostic views or mismeasuring anatomical structures~\cite{Sharma2021}. International guidelines underscore the critical role of \ac{us} in diagnosing fetal growth disorders, highlighting the need for standardized scanning protocols and quality control measures. However, intra- and inter-operator variability remains a major source of error~\cite{Salomon2022ISUOGScan,Sharma2021}. Thus, automated support tools are of great interest as they can help reduce variability, improve efficiency, and offer training support for less experienced operators.
\begin{figure*}
    \centering
    \includegraphics[width=0.8\linewidth]{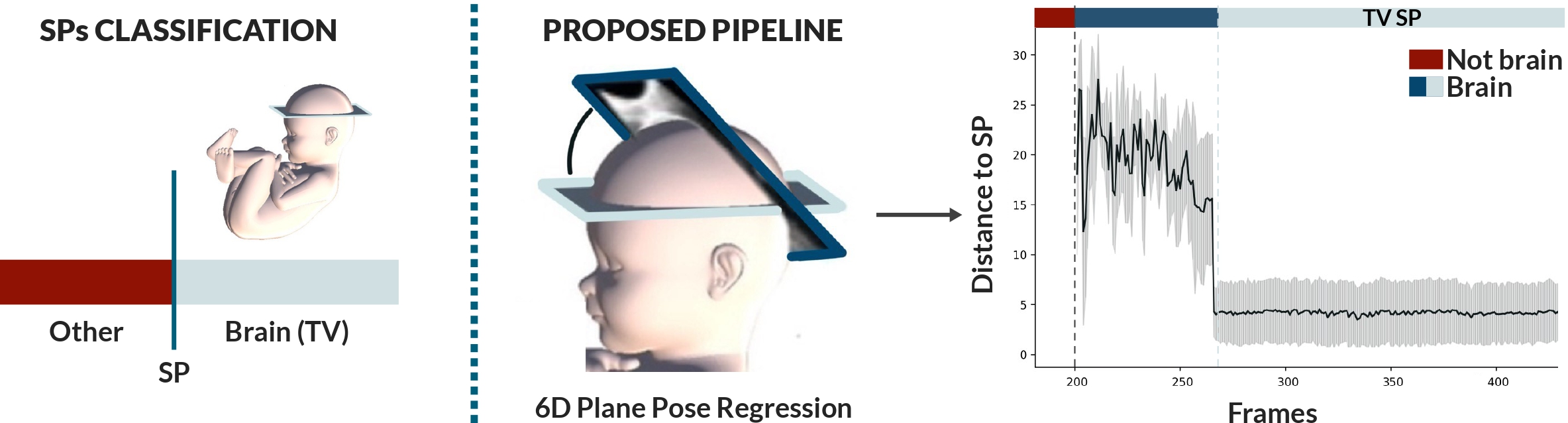}
   \caption{Comparison of classification-based approaches (left) that provide binary SP/non-SP labels and only signal when a good plane has been acquired with the proposed continuous proximity estimation approach (right) that outputs continuous proximity to the TV SP during scanning, driven primarily by translational distance (in millimeters) to provide a smooth, actionable trajectory, designed to enable earlier and more precise probe adjustments.}
    \label{fig:intro}
\end{figure*}

In this work, we introduce a pipeline that brings \ac{us} plane pose estimation closer to assisted navigation, focusing on the fetal brain in \ac{2d} \ac{us} video. Our system (Figure~\ref{fig:intro}) operates in three sequential stages: it detects when the fetal brain is visible, masks out irrelevant anatomy, and continuously predicts the proximity of the current frame to a specific target \ac{sp} (in this work, the \ac{tv} \ac{sp}). Masking out peripheral anatomy, such as maternal tissues, acoustic shadows, and amniotic fluid, ensures that the network extracts features solely from the fetal brain and prevents overfitting to background artifacts; the mask is internal to the pipeline and does not alter the full-field \ac{us} image on the machine's monitor, on which the sonographer relies for diagnosis. Unlike prior methods that assume the \ac{sp} is already acquired, our approach outputs a continuous proximity signal during scanning, designed to enable earlier and more precise adjustments. To provide intuitive feedback, the system isolates translational distance (in millimeters) as the primary guidance signal; because rapid wrist adjustments cause rotational deviation to fluctuate heavily, rotation is treated as a secondary background constraint rather than fused into a single, noisy score. This passive scalar metric does not provide explicit \ac{6d} directional vectors: it serves as an intuitive ``hotter/colder'' signal that tells the operator whether the current sweep is converging toward or diverging from the target, but not which direction to move the probe. It is therefore not proposed as the final guidance solution but as a step toward low-cognitive-load active guidance interfaces that map proximity to explicit directional cues.

Our contributions can be summarized as follows: 
1) We propose a semi-supervised classification and segmentation model (\emph{Class+SS-Seg}) that reduces the annotation burden by training on a limited set of labeled \acp{sp} and a large corpus of unlabeled slices from \ac{3d} \ac{us} volumes; it generalizes to arbitrary non-\acp{sp} of the fetal brain, where fully supervised models typically collapse;
2) We combine this model with a \ac{6d} plane pose regression network, showing that isolating the fetal brain anatomy improves pose regression accuracy over an unmasked-input baseline;
3) We formulate a continuous proximity-based metric for \ac{sp} navigation: using canonical \ac{tv} \ac{sp} poses annotated on \ac{3d} volumes, the pose regression network yields the relative spatial distance of arbitrary \ac{2d} video frames at inference time, transforming plane localization from a binary ``detected/not detected'' task into a continuous geometric trajectory;
4) We evaluate our pipeline on freehand \ac{us} videos of real fetal scans from sonographers with varying expertise. Because the evaluation is retrospective, we cannot yet measure absolute gains in navigation speed; however, operators naturally freeze near local proximity minima, and while continuous proximity is a geometric prerequisite for acquisition, it does not act as a standalone proxy for expert quality scores, underscoring the need to combine spatial proximity with visual landmark assessment;
5) We validate real-time, edge-device deployment capabilities (39~Hz).

\section{Related Work}
\label{sec:relwork}

A wide range of \ac{ai}-based approaches support fetal \ac{us} scanning, aiming to reduce inter-operator variability and improve the reliability of \ac{sp} acquisition. Existing methods span segmentation as a pre-processing step to enhance downstream tasks, probe guidance through pose estimation, similarity-based retrieval of frames or clips, temporal tracking across freehand \ac{us} videos and direct regression of the \ac{6d} plane orientation. Below, we review relevant methods and highlight their limitations in relation to our proposed approach.

Segmentation has often been used as a pre-processing step to improve downstream tasks such as classification or pose estimation. Yeung et al.~\cite{Yeung2021b,Yeung2022} integrated a segmentation network into their plane localization pipeline, ensuring that pose regression focused on the fetal brain region. Classifier networks have also been used to recognize \acp{sp} in real time~\cite{Baumgartner2017a}, and when combined with segmentation, can automate biometry on those planes~\cite{AutoFB,Avisdris2022BiometryNet:Planes}. These strategies demonstrate the value of filtering out irrelevant anatomy, but they typically rely on large, fully annotated datasets that are costly to produce and assume the \ac{sp} has already been acquired. 

Guidance systems aim to help less experienced sonographers navigate to \acp{sp} by estimating the pose of the acquisition plane. 
Traditional methods extracted \acp{sp} from pre-acquired \ac{3d} \ac{us} volumes~\cite{Rahmatullah2011,Cuingnet2013}, which support offline analysis but cannot guide scanning in real time. To overcome this, sensor-based solutions attach hardware to the probe: Droste et al.~\cite{Droste2020a} combined \ac{us} video with inertial sensor data in their US-GuideNet system to predict optimal probe movements, and Birlo et al.~\cite{Birlo2022} developed CAL-Tutor, an augmented reality-based training system overlaying target plane orientation cues via a HoloLens headset. While effective, such approaches require specialized equipment not routinely available in clinics. On the other hand, sensorless methods estimate probe pose directly from \ac{us} images. For instance, Pose-GuideNet~\cite{Men2024Pose-guidenet:Estimation} aligns freehand frames of the fetal brain to a \ac{3d} atlas and outputs movement directions. However, while these approaches remove hardware dependencies, they have been validated mainly on curated datasets. They still struggle with off-plane variability and with communicating pose information in a clinically intuitive way. 

Similarity-based approaches identify frames or clips visually resembling a target \ac{sp}. Gao et al.~\cite{Gao2020LabelLearning} used metric learning to embed \ac{us} images into a space where \acp{sp} form distinct clusters. Similarly, Alasmawi et al.~\cite{Alasmawi2024FUSC:Learning} performed self-supervised clustering of large, unlabeled second-trimester \ac{us} datasets into anatomical view categories, achieving high clustering purity without manual labels. Extending this idea to videos, Mishra et al.~\cite{Mishra2025MCAT:Transformer} framed visual query-based clip localization as a multi-tier, class-aware transformer to retrieve the segment containing the target anatomy, and TIER-LOC~\cite{Mishra2025TIER-LOC:Transformer} enhanced query-conditioned localization with hierarchical feature modeling to improve temporal coherence in long freehand scans. More recently, the application of vision-language models has shown promise in providing multimodal, action-oriented scanning assistance~\cite{guo2026visually}. While these methods offer highly flexible guidance and reduce annotation requirements, they typically provide broad navigational prompts or a binary decision (``\ac{sp}'' or ``not-\ac{sp}'') and do not quantify exactly how far the current plane lies from the target in \ac{6d} geometric space.

Temporal models leverage the continuity in \ac{us} videos to stabilize predictions and improve \ac{sp} detection. Chen et al.~\cite{Chen2015AutomaticNetworks} pioneered the use of recurrent neural networks for \ac{sp} detection in streaming data. P\l{}otka et al.~\cite{Potka2021FetalNet:Measurements,Potka2022DeepMeasurements} proposed \textit{FetalNet}, a multi-task framework coupling plane detection with biometric measurement directly from freehand video, and demonstrated expert-level agreement in fetal biometry on large datasets. These methods demonstrate that video-based tracking can reduce false detections and produce clinically valid measurements, but they primarily confirm or refine \acp{sp} after they have already been acquired, rather than guiding the operator toward them. 


Pose regression has emerged as a promising approach for estimating the \ac{6d} pose of \ac{us} planes from a single \ac{2d} image, enabling sensorless guidance. Early volumetric methods cast \ac{sp} localization as an iterative transform prediction problem in \ac{3d} volumes, where a \ac{cnn} estimates the motion needed to move a current slice toward the target plane, repeating until convergence~\cite{Li2018}. Di Vece et al.~\cite{DiVece2022} later proposed a \ac{cnn}-based regression of fetal brain plane orientation directly from \ac{2d} volume slices. Similarly, Yeung et al.~\cite{Yeung2021b} introduced self-supervised strategies to map \ac{2d} frames into a \ac{3d} fetal atlas, and later fine-tuned these models on real freehand \ac{us} videos to improve generalization~\cite{Yeung2022}. Building on these advances, our prior work defined a generalized coordinate system to make \ac{6d} predictions more consistent across fetuses~\cite{DiVece2023}. Most recently, Dou et al.~\cite{Dou2025StandardGuidance} have proposed a probabilistic approach using denoising diffusion with multi-scale guidance and a spherical plane parameterization to improve angular sensitivity across multiple \acp{sp}. While effective, these methods have been validated mainly on synthetic slices, highly curated datasets, or static \ac{3d} volumes, and their outputs (\ac{6d} pose vectors or plane parameters) are not directly interpretable for clinical navigation. 

While classifiers, segmentation networks, retrieval strategies, and temporal models have improved recognition and measurement on \acp{sp}, they generally assume that the correct plane has already been acquired. Sensor-based and pose regression methods move closer to real-time guidance but often require curated training data or volumetric input or produce outputs that are not directly interpretable. Our work instead couples semi-supervised segmentation with pose regression to deliver a continuous, operator-oriented proximity measure during freehand scanning.

\section{Methods}
\label{sec:methods}

Our proposed pipeline involves three main steps: i) classification and segmentation (\emph{Class+SS-Seg}) of the fetal brain from a \ac{us} image; ii) plane pose regression on masked fetal brain images to estimate \ac{6d} plane pose in normalized space; iii) measurement of proximity using Euclidean distance (translation) to the target \ac{sp} and the angle between plane normals (rotation). The method is trained in a semi-supervised manner using a combination of labeled \ac{sp} images and unlabeled images sliced from \ac{3d} \ac{us} volumes (non-\acp{sp}).

\subsection{Semi-Supervised Fetal Brain Segmentation}
\label{methods:network:ss}

\begin{figure*}
    \centering
    \includegraphics[width=0.9\linewidth]{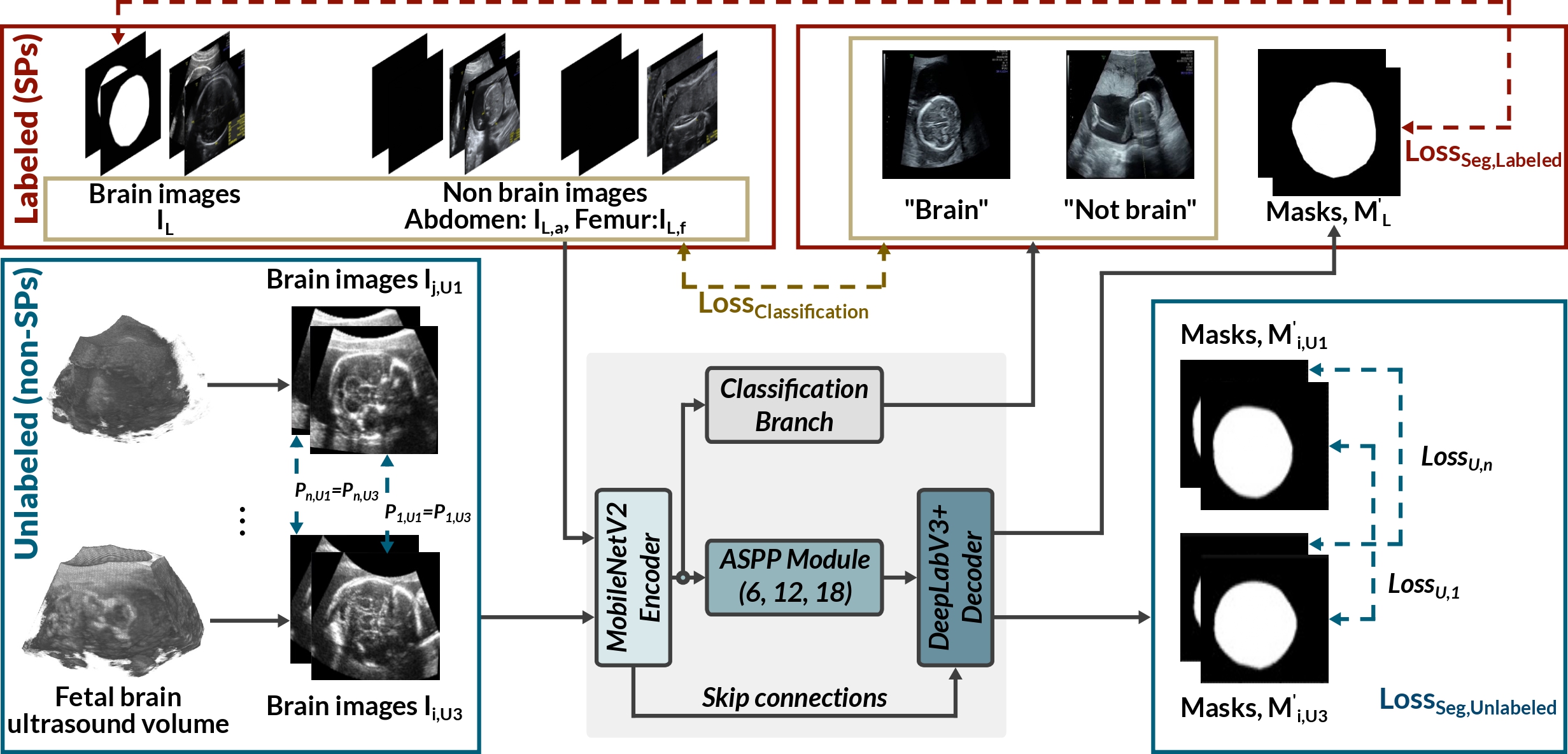}
    \caption{Semi-supervised model: MobileNetV2 encoder (pretrained on ImageNet), Atrous Spatial Pyramid Pooling  (dilations 6, 12, 18), decoder with skip connections for mask prediction, and classification branch for brain/non-brain discrimination. Labeled images (SPs: brain and non-brain) are supervised with Dice and weighted binary cross-entropy (Loss$_{\text{Seg,lab}}$); for unlabeled non-SP slices from different volumes at the same pose, a consistency loss (Loss$_{\text{Seg,unlab}}$) enforces similar predictions across volumes.}
    \label{fig:architecture}
\end{figure*}

Existing labeled datasets for fetal brain segmentation include only \acp{sp}, and manually annotating a large, diverse set of arbitrary non-SPs is prohibitively expensive and labor-intensive. We therefore formulate a semi-supervised learning approach that leverages a small set of fully labeled \ac{2d} \ac{sp} images alongside a large corpus of unlabeled non-\ac{sp} slices rendered directly from \ac{3d} \ac{us} volumes.
Our \emph{Class+SS-Seg} network (Figure~\ref{fig:architecture}) is built on the DeepLabV3+~\cite{deeplabv3+} architecture with MobileNetV2 pretrained on ImageNet.
The encoder takes as input an image with dimensions C$\times$H$\times$W, where $C=3$ is the number of channels and $H=W=320$ the height and width, and generates a shared feature representation, which is passed to the segmentation decoder and classification head. DeepLabV3+ employs \ac{aspp} to capture multi-scale context, which we found beneficial for \ac{us} segmentation. The network is modified to produce two outputs: a binary classification indicating whether the frame contains fetal brain tissue, and a segmentation mask. We implement the classification branch as an auxiliary output in the DeepLabV3+ model, using global average pooling followed by a fully connected layer and a sigmoid activation to produce a ``brain versus not brain'' probability. The segmentation head produces a single-channel probability map for the fetal brain region, whether the plane is an exact \ac{sp} or an arbitrary slice. We train our model using a combination of supervised and consistency losses. The classification branch is trained in a fully supervised manner using \ac{BCE} ($\mathcal{L}_{\text{Class}}$). Images containing the fetal brain serve as positive examples, while images of other anatomies (``not brain'' in Figure~\ref{fig:architecture}) serve as negative examples, for which an empty (black) mask is used as the ground truth. 

The segmentation branch is trained using a compound loss on labeled data and a consistency loss on unlabeled data. On labeled images with ground-truth mask $M$, we minimize a weighted sum of Dice loss and \ac{wBCE} loss. The latter is used to address class imbalance, since background pixels far outnumber brain pixels. We compute the weights from the inverse pixel frequencies in the training masks.
We define the weight for the brain class as $w_1~=~\frac{\max(N_0, N_1)}{N_1}$ and for background as $w_0~=~\frac{\max(N_0, N_1)}{N_0}$. In practice, $N_0 > N_1$, so $w_1~=~N_0/N_1$ (a number greater than 1) and $w_0~=~1.0$. The supervised segmentation loss on labeled data (\acp{sp}) is then:
\begin{equation}
    \mathcal{L}_{\text{Seg,lab}}~=~\frac{\mathcal{L}_{\text{Dice}}(M'_L, M) + \mathcal{L}_{\text{wBCE}}(M'_L, M)}{2}
\end{equation} 
where $M'_L$ is the predicted mask for a labeled image and $M$ is the ground-truth mask. Combining Dice ($\mathcal{L}_{\text{Dice}}$) and \ac{wBCE} ($\mathcal{L}_{\text{wBCE}}$) losses is a well-established practice in medical image segmentation that balances structural overlap with attention to small, under-represented structures.

For the unlabeled brain images (non-\acp{sp}), a consistency loss $\mathcal{L}_{\text{Seg,unlab}}$ enforces similar segmentations for slices extracted at identical relative poses across registered volumes. 
Given a pair (or group) of unlabeled images acquired at the same pose from different volumes, we encourage the model to output the same mask for all images in the pair (or group). 
Let $M'_{U,i}$ and $M'_{U,j}$ be the predicted masks (or probability maps) for two unlabeled images $I^i_U$ and $I^j_U$ taken at the same pose (from volumes $i$ and $j$, respectively). We define a consistency loss between this pair of predictions as the average of the Dice and \ac{wBCE} losses; this penalizes differences in both overlap and pixel-wise probabilities. In practice, to stabilize training, we compute the Dice consistency on the probability maps (after sigmoid) and the \ac{wBCE} consistency on the logits in a symmetric fashion (using $M'_{U,i}$'s logit against $M'_{U,j}$'s probability and vice versa). For a group of $n$ volumes, we consider all $\binom{n}{2}$ pairs. For example, with $n=3$ volumes per group, we have three pairs $(1,2), (1,3), (2,3)$; we compute the consistency loss for each pair and average them. The unlabeled segmentation loss can be written as:
\begin{equation}
\begin{aligned}
\mathcal{L}_{\text{Seg,unlab}}
= \frac{1}{|P|}\sum_{(i,j)\in P}
\left[
\begin{aligned}
& \mathcal{L}_{\text{Dice}}\!\left(M'_{U,i},\, M'_{U,j}\right) \\
& + \mathcal{L}_{\text{wBCE}}\!\left(M'_{U,i},\, M'_{U,j}\right)
\end{aligned}
\right]
\end{aligned}
\end{equation}
where $P$ is the set of volumes in the group (P=3 in the training set and P=2 in the validation set). The loss term drives the predictions $M'_{U,i}$ and $M'_{U,j}$ to be as similar as possible for images that show the same anatomical cross-section across different fetuses. By minimizing $\mathcal{L}_{\text{Seg,unlab}}$, the model learns shape consistency and generalizes better to arbitrary planes that are not in the labeled set. The total loss is then a sum of all three components:
\begin{equation}
    \mathcal{L}_{\text{Tot,Class+SS-Seg}} = \mathcal{L}_{\text{Seg,lab}} + \alpha \mathcal{L}_{\text{Seg,unlab}} + \beta \mathcal{L}_{\text{Class}}
\end{equation}
where $\alpha=0.5$ weights the unlabeled consistency term at half the supervised segmentation term and $\beta=1.0$ weights the classification loss. The classification loss $\mathcal{L}_{\text{Class}}$ is only applied to images from the labeled set (\acp{sp} with a known ``brain/not brain'' label). In the early training epochs, we found it beneficial to focus on segmentation first and set $\beta=0$ (no classification loss; model \emph{SS-Seg}) to allow the encoder and decoder to learn robust segmentation features, even on very challenging slices, without distraction from classification. Subsequently, we fine-tune the model by enabling the classification branch ($\beta=1$, model \emph{Class+SS-Seg}) and continuing training with a lower learning rate. This staged training prevents the classification task from interfering with early-stage segmentation learning on limited labeled data, while still allowing the encoder to benefit from the auxiliary signal in later training stages.

\subsection{Plane Pose Regression}
\label{methods:network:posereg}

We use an 18-layer residual \ac{cnn} (ResNet-18)~\cite{He2016} as a backbone for feature extraction with pretrained ImageNet weights. We modify the network by reinitializing the fully connected layer to match the representation's dimension (9 parameters) and adding a regression head to output the rotation and translation representations. The network receives a masked \ac{us} image $I$ (256$\times$256) obtained by slicing the volume and applying the mask generated by the \emph{Class+SS-Seg} network, and its \ac{6d} pose with respect to the brain center of the registered volumes~\cite{DiVece2023} $\mathbf{p_{GT}}=(t_x, t_y, t_z, \alpha_x, \alpha_y, \alpha_z)$. We use this information as the ground-truth label for network training and validation. The \ac{cnn} learns to predict the \ac{6d} pose $\mathbf{p_{Pred}}=(t'_x, t'_y, t'_z, \alpha'_x, \alpha'_y, \alpha'_z)$. 

For both translation and rotation, we use the \ac{mse} between predicted ($\mathbf{t}'$,$\mathbf{R}'$) and ground-truth ($\mathbf{t}$,$\mathbf{R}$) values: 
\begin{equation}
\mathcal{L}_{\text{Transl}} = \frac{1}{N}\sum_{i=1}^{N}\left \| \mathbf{t}'-\mathbf{t} \right \|_2^2, \mathcal{L}_{\text{Rot}} = \frac{1}{N}\sum_{i=1}^{N}\left \| \mathbf{R}'-\mathbf{R} \right \|_F^2
\end{equation}
where $N$ denotes the total number of images $I$ within one training epoch, $\mathbf{t}'$ denotes the predicted translation component and $\mathbf{t}$ the ground-truth label. $\mathbf{R}$ is the 3$\times$3 rotation matrix obtained from the ground-truth Euler angles $\mathbf{r}~=~(\alpha_x,\alpha_y,\alpha_z)$ (in radians) and $\mathbf{R}'$ is the 3$\times$3 rotation matrix obtained from the six parameters $r_1,\dots,r_6$ as the output of the network, providing a smooth and differentiable representation for any arbitrary \ac{3d} rotation.
The total loss function is then computed as a fixed weighted sum:
\begin{equation}
\mathcal{L}_{\text{Tot}}~=~(1-\lambda)\mathcal{L}_{\text{Rot}} + \lambda \mathcal{L}_{\text{Transl}}
\end{equation}
where $\lambda$ is a balancing coefficient. However, because the relative scale and difficulty of the two tasks differ, we employ an uncertainty-weighted loss $\mathcal{L}_{\mathcal{U}}$ based on the homoscedastic uncertainty formulation for multi-task learning. We introduce two trainable scalar parameters $s_r$ and $s_t$, which parameterize the (log) task uncertainties, defining the total loss as:
\begin{equation}
\mathcal{L}_{\mathcal{U}} =
\frac{1}{2\,\exp(s_r)}\,\mathcal{L}_{\text{Rot}} 
+ \frac{1}{2\,\exp(s_t)}\,\mathcal{L}_{\text{Transl}} 
+ \frac{1}{2}\left(s_r+s_t\right)
\end{equation}
In practice, $s_r$ and $s_t$ are initialized to zero and optimized jointly with the network weights. Tasks with higher residual error (\emph{i.e.,} larger effective uncertainty) are automatically down-weighted, whereas more reliable tasks are emphasized, eliminating the need to tune $\lambda$ manually. 

For pose regression experiments on volume slices (\ac{loocv}), we report the error as Euclidean distance for translation and the $SO(3)$ geodesic angle between predicted and ground-truth rotation matrices (in degrees) for rotation:
\begin{equation}
\alpha_{rot}^{\mathrm{geo}}~=~\arccos\left(\frac{\mathrm{trace}(R_{\text{pred}}R_{\text{gt}}^{\top})-1}{2}\right)
\end{equation}
For proximity estimation on \ac{us} videos, we compute translational and rotational distances to the target separately during inference. We use Euclidean distance for translation (in-plane and out-of-plane components combined):
\begin{equation}
d_{trans}~=~\left\| \mathbf{t_{frame}} - \mathbf{t_{SP}} \right\|_2
\end{equation}
For rotation, we measure the angle between the normal vectors of the current frame ($\mathbf{N}_{frame}$) and the target \ac{sp} ($\mathbf{N}_{SP}$):
\begin{equation}
\alpha_{rot}^{\mathrm{n}}~=~\arccos\left( \frac{\mathbf{{N}_{frame}} \cdot \mathbf{{N}_{SP}}}{\| \mathbf{{N}_{frame}} \| \| \mathbf{{N}_{SP}} \|} \right)
\end{equation}
This formulation excludes in-plane rotation (about the plane normal): a version of the \ac{sp} rotated in-plane is coplanar with the target, has $\alpha_{rot}^{\mathrm{n}}~=~0$, and remains clinically valid because in-plane rotation does not affect biometric measurements (\emph{e.g.,} head circumference, biparietal diameter).
However, we account for in-plane translation, as shifting the imaging plane laterally within the target \ac{sp} can move anatomical landmarks out of view, thereby affecting measurement accuracy. For example, a plane parallel to the \ac{tv} \ac{sp} but displaced by 5\,mm may miss key structures. Thus, $d_{trans}$ captures both out-of-plane displacement (perpendicular to the \ac{sp}) and in-plane displacement (parallel shift within the plane). Decoupling the two metrics prevents high-frequency angular noise from obscuring the operator's translational trajectory and lets the system prioritize a smooth, actionable translational signal while monitoring rotational alignment in the background.

\section{Experiments and Results}
\label{sec:exp-results}

All models are implemented in \emph{PyTorch} and trained using a single Tesla\textregistered{} NVIDIA A100-SXM4-40GB GPU.


\subsection{Datasets}
\label{methods:dataset}

\begin{figure*}
    \centering
    \includegraphics[width=\linewidth]{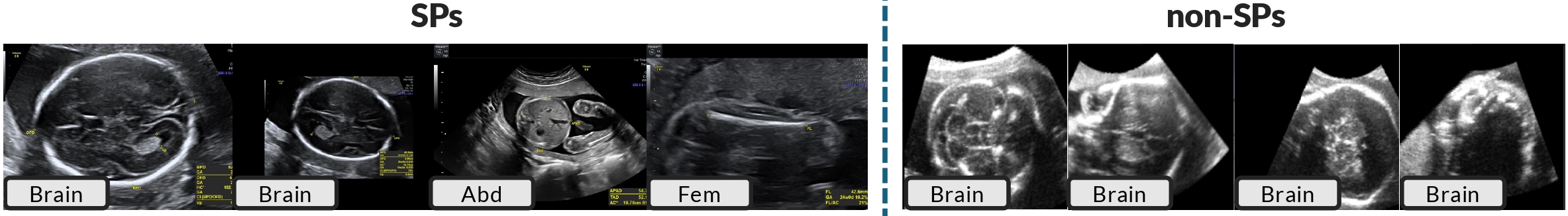}
    \caption{Labeled standard planes (SPs) used for training (left: brain, with and without black borders, abdomen, and femur) versus non-standard planes (non-SPs, right), whose variable, partial cross-sections motivate semi-supervised segmentation beyond SPs.}
    \label{fig:plane-comparison}
\end{figure*}

To train and validate the subcomponents of our pipeline, we use a combination of images sliced from \ac{3d} \ac{us} volumes (unlabeled non-\acp{sp}) and a \ac{2d} \ac{us} dataset (which provides labeled \acp{sp}). A visual comparison of the \acp{sp} and non-\acp{sp} is shown in Figure~\ref{fig:plane-comparison}. Finally, to validate the clinical applicability of the entire pipeline, we evaluate our system on an in-house dataset of full-length, freehand clinical \ac{us} videos.  
\subsubsection{Slice from 3D Ultrasound Volumes (Unlabeled non-SPs)} We use six fetal brain \ac{us} volumes from~\cite{Pistorius2010} (singleton pregnancies with no abnormal findings) acquired at the axial \ac{tv} \ac{sp} from pregnancies at 21--25 weeks \ac{ga} (two volumes at 23 weeks), aligned and processed to an isotropic voxel size of 0.5\,mm and an average size of 249$\times$174$\times$155\,mm~\cite{DiVece2023}. We slice these volumes on a deterministic \ac{6d} grid of rotations and offsets around the volume center, bounded to avoid slices with poor overlap with the volume~\cite{DiVece2023}. We extract 22,029 images per volume. To evaluate segmentation performance on these non-\acp{sp}, an experienced obstetrician annotated the binary masks of 10 randomly selected images per volume using the VIA tool~\cite{Dutta2016VGGVIA}. Additionally, the obstetrician annotated the canonical pose of the \ac{tv} \ac{sp} on these volumes using a \ac{3d} interface~\cite{DiVece2023} to provide ground-truth targets for pose regression and proximity estimation.

\subsubsection{2D Ultrasound Images (Labeled SPs)} In addition to volume slices, we use a \ac{us} dataset of 346 \ac{us} images acquired from 42 fetuses during the second trimester with labeled segmentation masks of the brain, abdomen, and femur initially used in the AutoFB work~\cite{AutoFB} and then expanded for the dataset detailed in~\cite{divece2025}. This dataset provides the expertly labeled ground-truth masks needed to train and test the semi-supervised models: 135 brain images and 211 non-brain images (abdomen and femur). To expose the classifier to the cropped, black-bordered framing seen on clinical machines, the brain images are additionally included with synthetic black borders (9 images), bringing the labeled set to 355 images in total. 
\subsubsection{Clinical Ultrasound Videos (Pipeline Validation)} Finally, we test the complete pipeline on 17 routine second-trimester fetal \ac{us} scans recorded at the \ac{us} Screening Unit of University College London Hospital \ac{nhs} Foundation Trust (prospective observational study between February and June 2023, approved by the \ac{nhs} Research Ethics Committee~\cite{lelous2024probe}). Scans were acquired on a Voluson\textsuperscript{TM} E8\textregistered{} \ac{us} machine (GE Healthcare, Chicago, Illinois) by sonographers of varying expertise, 14 senior \ac{us} fellows (novices, $<$5 years' experience) and 3 fetal medicine consultants (experts, $>$5 years' experience), following routine mid-trimester anomaly scanning protocols, with informed consent from each participant. The 17 scans were recorded by 17 distinct operators and encompass diverse fetal presentations (cephalic and breech) and maternal body mass indices (mean $\pm$ standard deviation: 26.9 $\pm$ 6.3 $\mathrm{kg/m^2}$). Mean scan duration was approximately 19--21 minutes, with substantial variability in scanning workflow and probe motion. The full videos used in our pipeline will be made publicly available.
\subsection{Brain Classification and Segmentation} 
\label{exp-results:semanticsegmentation}

The labeled images were split into 205 for training, 53 for validation, and 97 for testing, with a balanced distribution of ``brain'' and ``not brain'' (abdomen and femur) classes in each set. Images are cropped and resized to 320$\times$320. In our semi-supervised approach, only the ``brain'' class carries a non-empty mask; abdomen and femur images enter the segmentation loss with empty masks and serve as negative examples for the classifier. The unlabeled volume slices, in turn, are split into six folds (one for each volume). To evaluate generalization and avoid any overlap between the training and test data, we use \ac{loocv} across the six sets of unlabeled images. In each of the six folds, we designate one volume as ``test'' for downstream pose regression, hold out another two volumes for unlabeled validation, and use the remaining three volumes for unlabeled training. The fixed set of 205 labeled \ac{2d} images is used for training in every fold (and the 53 labeled validation images for validation in every fold), combined with the unlabeled slices from that fold's training volumes. This means we train six segmentation models, each time using a distinct subset of unlabeled data, and evaluate each model on its held-out volume and on the common labeled test set.

Each segmentation model is trained in three stages. First, we train a model \emph{S-Seg} in a fully supervised manner on the fixed labeled training set using a learning rate $lr=0.001$; this corresponds to $\alpha=0$ and $\beta=0$ in our loss. Flipping, rotation, Gaussian noise, elastic transformation, and brightness and contrast augmentations are applied. Then we add the unlabeled training set and obtain the model \emph{SS-Seg} ($\beta=0$) by training with a learning rate of $lr=0.008$. Finally, we add the classification head and train the model \emph{Class+SS-Seg} with $lr~=~0.00003$. We use a batch size of 8 during the entire process and the Adam optimizer. Rather than setting a fixed number of epochs per stage, we employ an early-stopping strategy based on validation loss. We train for up to 600 epochs and stop if the validation loss does not improve for 100 epochs. The much lower learning rate in the final stage stabilizes training and prevents disruption of the segmentation features already learned (Section~\ref{methods:network:ss}).
 
We compare our results against AutoFB~\cite{AutoFB}, which performs 3-class segmentation of the femur, brain, and abdomen. It was trained in a fully supervised manner on the same labeled training set as our model for 600 epochs, with an early-stopping patience of 50 epochs. For a fair comparison, we retrain the model using black masks for the abdomen and femur, as done in our experiments.
For comparison with prior architectures, we train a U-Net~\cite{unet2015} + MobileNetV2 model under the same protocol: this model uses the same training data and losses, but with a U-Net decoder (and no \ac{aspp}) instead of DeepLabV3+. Table~\ref{tab:iou-res} reports the mean \ac{iou} (mIoU) for the labeled test set (\acp{sp} in Figure~\ref{fig:plane-comparison}). We also report the m\ac{iou} for the randomly selected subset of 10 non-\acp{sp} from each test volume (Figure~\ref{fig:plane-comparison}; Section~\ref{methods:dataset}). For \emph{SS-Seg} and \emph{Class+SS-Seg}, we report the average performance of our six trained models with the \ac{loocv}. For \emph{S-Seg} and AutoFB, we report only one model, trained on the fixed labeled training set and tested on the same six test volumes used in the \ac{loocv} (average). As summarized in Table~\ref{tab:iou-res}, the supervised \emph{S-Seg} and AutoFB baselines segment \acp{sp} accurately but collapse on non-\acp{sp}, whereas \emph{SS-Seg} and especially \emph{Class+SS-Seg} retain strong performance on both.
\begin{table}
\centering
\caption{Mean Intersection over Union (mIoU) on \acp{sp} and non-\acp{sp} (LOOCV) for \mbox{AutoFB~\cite{AutoFB}}, \emph{S-Seg}, \emph{SS-Seg}, and \emph{Class+SS-Seg} (DeepLabV3+ and UNet). \textbf{Bold}: best; \underline{underline}: second-best.}
\label{tab:iou-res}
\resizebox{\columnwidth}{!}{%
\begin{tabular}{cccccc} 
\toprule
\textbf{mIoU} &\textbf{Architecture} &\textbf{AutoFB~\cite{AutoFB}} &\textbf{S-Seg} & \textbf{SS-Seg} & \textbf{Class$+$SS-Seg} \\ 
\midrule
\multirow{2}{*}{SPs} & DeepLabV3+ & \textbf{0.9695}& \underline{0.9596}& 0.8570& 0.9282\\
 & UNet & \multicolumn{1}{c}{\textbf{0.9705}} & \multicolumn{1}{c}{\underline{0.9627}} & \multicolumn{1}{c}{0.8632} & \multicolumn{1}{c}{0.9301} \\ \midrule
\multirow{2}{*}{non-SPs} & DeepLabV3+ & 0.1003& 0.4084& \underline{0.8123}& \textbf{0.8613}\\
 & UNet & \multicolumn{1}{c}{0.1333} & \multicolumn{1}{c}{0.6813} & \multicolumn{1}{c}{\underline{0.7640}} & \multicolumn{1}{c}{\textbf{0.8334}} 
 \\
\bottomrule
\end{tabular}
}
\end{table}
\subsection{Plane Pose Regression}
\label{results:poseregression}
For both training and inference with our pose regression model, we begin by applying the brain segmentation predictions, dilating them with a 30$\times$30 kernel, and masking out irrelevant regions of the image. We train six pose regression models on masked images (one per segmentation model) for \ac{loocv}. We use 20\% of the training data for validation. The network is trained for 200 epochs with a batch size of 64 using the Adam optimizer. 
Table~\ref{tab:posereg-loocv-res} summarizes \ac{loocv} performance across configurations applying segmentation masks and shows that masking improves pose estimation over unmasked input. 

During real-time freehand scanning, the displayed \ac{us} frames can appear horizontally flipped depending on the probe orientation and the scanning direction. Since our pose regression model is trained on slices extracted from registered \ac{3d} volumes with a fixed canonical orientation, this mismatch can degrade performance when deploying on clinical videos. To improve robustness to such appearance changes, we introduce horizontal flip augmentation during training: with probability 0.5, each input image is horizontally flipped, while its corresponding pose labels remain unchanged. Because pose is defined in the canonical volume coordinate system rather than the image coordinate system, this augmentation preserves the underlying \ac{3d} plane (and thus its ground-truth pose) while exposing the network to mirrored image appearances encountered in real scans. We report this as the ``flip'' configuration in Table~\ref{tab:posereg-loocv-res}. In the baseline setting, we combine the rotation and translation losses with a fixed weighting factor $\lambda$~=~0.01. Across the six \ac{loocv} models, the uncertainty-weighted formulation (Section~\ref{methods:network:posereg}) converged to stable task weights (translation:rotation ratio $0.26\pm0.06$), producing the ``uncertainty'' configuration reported in Table~\ref{tab:posereg-loocv-res}. Notably, while flip augmentation degrades \ac{loocv} performance on the controlled volume slices (and flipping the pose label along with the image does not resolve it), this augmentation proved essential for successful deployment on real clinical videos (Section~\ref{sec:discussion}). Uncertainty weighting partially recovers the translational accuracy lost to flip augmentation, though not rotation (Table~\ref{tab:posereg-loocv-res}).
\begin{table*}
\centering
\caption{Translation and rotation errors (median, mean, minimum, and maximum across 6-fold LOOCV) of the pose regression model with and without segmentation masking for each training configuration. \textbf{Bold}: best; \underline{underline}: second-best.}
\label{tab:posereg-loocv-res}
\begin{tabular}{lcccccccc}
\toprule
\textbf{Configuration} & \textbf{trans$_{median}$} & \textbf{trans$_{mean}$} & \textbf{trans$_{min}$} & \textbf{trans$_{max}$} & \textbf{rot$_{median}$} & \textbf{rot$_{mean}$} & \textbf{rot$_{min}$} & \textbf{rot$_{max}$}  \\ 
\midrule
No masks, $\lambda=0.01$, no flip 
    & 2.97 & 4.11 & 0.05 & 47.48 & \underline{6.63} & \underline{8.96} & \underline{0.16} & \underline{126.52} \\ 
With masks, $\lambda=0.01$, no flip 
    & \textbf{2.49} & \textbf{3.35} & \textbf{0.02} & \textbf{42.53}      & \textbf{5.62} & \textbf{6.96} & \textbf{0.09} & \textbf{109.32} \\
With masks, $\lambda=0.01$, flip 
    & 3.74 & 4.99 & 0.11 & 46.30 & 7.44 & 13.76 & 0.18 & 178.45 \\
With masks, uncertainty weighting, flip
    & \underline{2.66} & \underline{3.70} & \underline{0.04} & \underline{43.53} & 8.16 & 14.67 & 0.23 & 177.27 \\
\bottomrule
\end{tabular}
\end{table*}

\subsection{Ultrasound Videos}
\label{results:videos}
Two versions of the \ac{us} videos are released: full-examination recordings with interface elements and sonographer annotations, and shorter, brain-focused clips obtained via a manual spatial crop that removes static interface elements and non-imaging regions. This crop is applied consistently across all pipelines as a dataset-level preprocessing step.
\begin{figure*}
    \centering
    \includegraphics[width=0.9\linewidth]{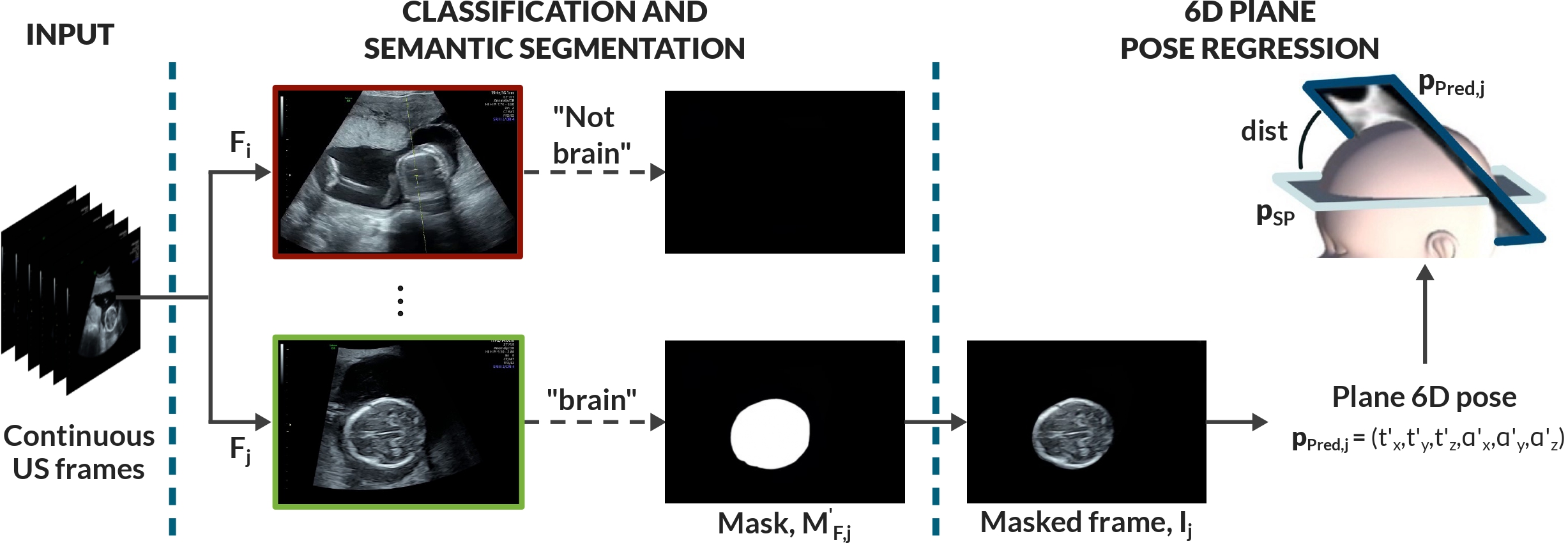}
    \caption{Inference pipeline on US video frames: frames classified as containing the brain are masked and fed to the pose regression network, which regresses the 6D pose and computes the distance to the TV SP.}
    \label{fig:inference-pipeline}
\end{figure*}

Frames extracted at 10~Hz are fed to the \emph{Class+SS-Seg} network to detect and mask the fetal brain; each masked frame is then passed to the flip-augmented, uncertainty-weighted pose regression network to compute distance to the pre-annotated \ac{tv} \ac{sp} in a canonical reference frame where the \ac{sp} has fixed and known coordinates (Figure~\ref{fig:inference-pipeline}) without further annotations at inference time. 
In parallel, we run the full video through a pretrained SonoNet-64 in its publicly available PyTorch implementation\footnote{\href{https://github.com/baumgach/SonoNet-weights}{\textcolor{blue}{https://github.com/baumgach/SonoNet-weights} (accessed August 2026)}}. 

SonoNet assumes the \ac{us} beam is centered and consistently scaled, but in freehand scans, zooming and probe repositioning cause substantial frame-to-frame changes, degrading its accuracy. For SonoNet inference only, we apply a dynamic crop tracking the beam over time: after the initial manual crop to remove user interface elements, we threshold each frame to obtain a binary mask of the active sector, take its largest connected component and tight bounding box, expand it by a fixed margin, temporally stabilize the box with an exponential moving average (falling back to the last valid window if detection fails), and resize to SonoNet's input resolution. As set out in Section~\ref{methods:network:posereg}, we treat translational proximity ($d_{trans}$) as the primary guidance signal and rotational proximity ($\alpha_{rot}^n$) as a secondary, background descriptor. 
\subsubsection*{Proximity Signal at Freeze Events}
\label{results:videos:freeze}

\begin{figure}
    \centering
    \includegraphics[width=\columnwidth]{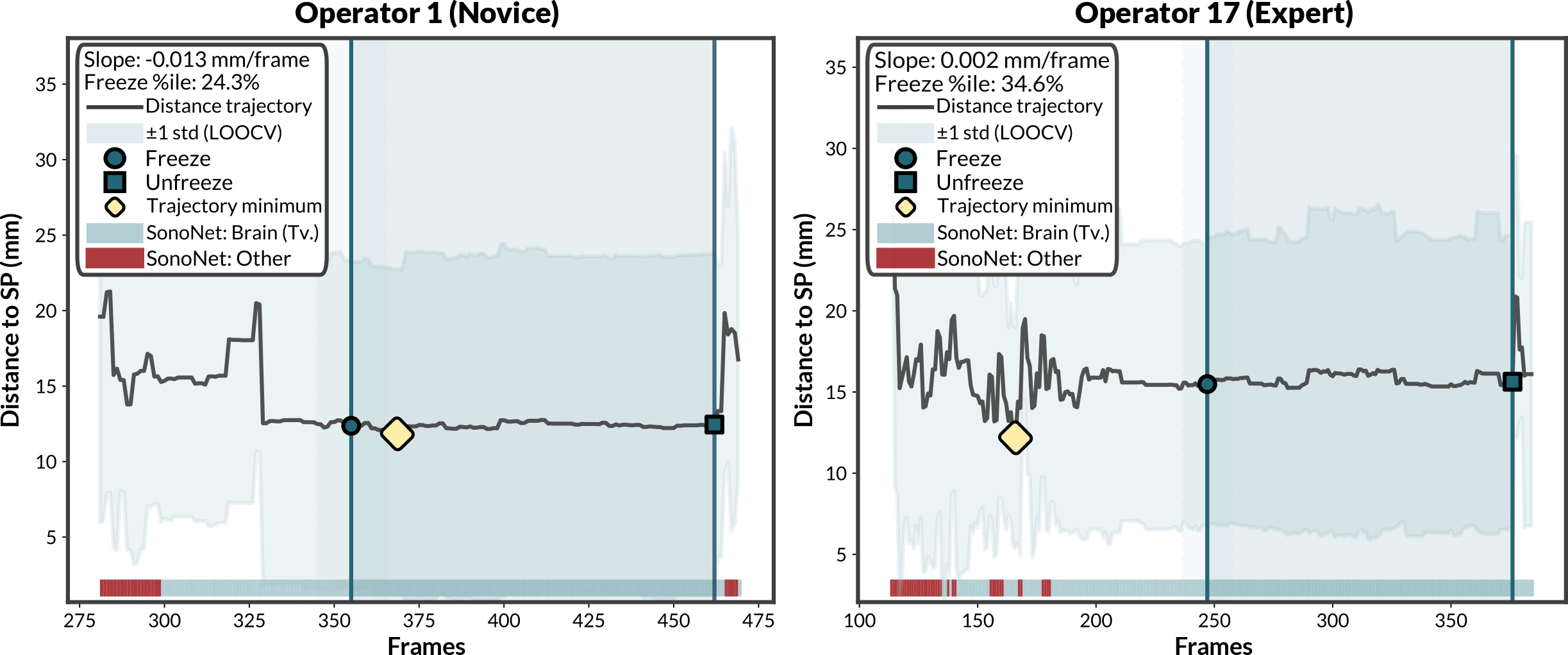}
    \caption{Novice (Operator 1) and expert (Operator 17) trajectories: translational distance to the TV SP over time (black; shaded band: $\pm1$ std across the six LOOCV models), $\pm1$\,s freeze window (light shading), freeze/unfreeze events (vertical lines), and trajectory minimum (diamond). The bottom line indicates SonoNet predictions. Freeze percentiles: 24.3\% (novice) and 34.6\% (expert).}
    \label{fig:example_traj}
\end{figure}

For each operator, we locate the freeze frame and compute its translational distance percentile within that operator's own trajectory. Because a manual freeze may lag the closest frame, we also take the best percentile within a $\pm$1\,s window (10\,Hz). Freezes landed near, but not exactly at, local trajectory minima: the best percentile within the window (mean 34.5\%) was significantly better than the exact-freeze percentile (mean 45.6\%; Wilcoxon signed-rank test, $n=17$, $W=136$, $p=0.0002$), and the window minimum fell in the bottom quartile of the trajectory for 8/17 operators. Freeze timing was nonetheless not driven by trajectory-level optimization: although most operators (11/17) froze below their trajectory-mean distance, the distance slope over the 2\,s before freeze was negative for only 8/17 operators and showed no significant approach trend ($t(16)=-0.903$, one-tailed $p=0.190$), the global-minimum frame preceded the freeze by a median of 10.7\,s, and cross-model agreement at freeze (standard deviation across the six \ac{loocv} models) was comparable to the trajectory average (ratio 1.03; lower at freeze for 8/17 operators) rather than enhanced. Representative expert and novice trajectories, showing the freeze frame, the $\pm1$\,s window, and the window and global minima, together with the SonoNet classification, are given in Figure~\ref{fig:example_traj}. 

\subsubsection*{Quality Scores, Operator Experience, and Proximity Metrics}
\label{results:videos:quality}

\begin{figure}
    \centering
    \includegraphics[width=\linewidth]{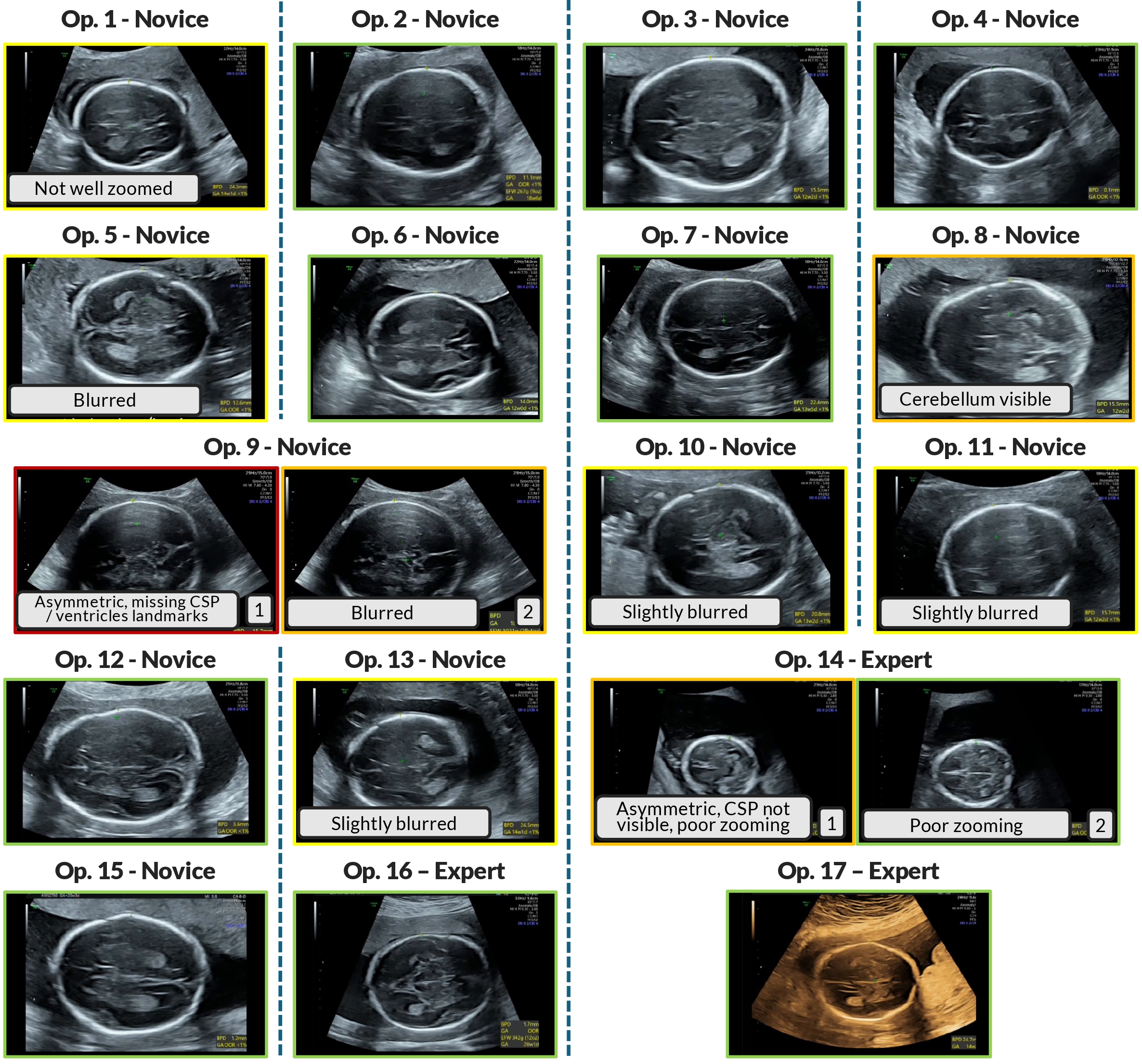}
    \caption{Visualization of TV SPs captured during live image acquisition for each operator. Image quality was assessed by an experienced obstetrician using a 5-point scale: green borders indicate a score of 5; orange, 4.5; yellow, 4; and red, 3.}
    \label{fig:freeze-tv-sp}
\end{figure}

The clinical quality of each operator's frozen \ac{tv} \ac{sp} was assessed using a 5-point scale (1~=~poor, 5~=~excellent) based on anatomical plane adequacy, zooming, and image clarity~\cite{Salomon2022ISUOGScan}. The assessment was performed by the same experienced obstetrician who annotated the canonical \ac{tv} \ac{sp} poses (Section~\ref{methods:dataset}); this reviewer was not one of the 17 operators who performed the recorded scans and was blinded to their experience levels. The heterogeneity of frozen plane appearance across operators and the anatomical or acquisition-related factors associated with lower scores are reflected in the per-operator assessments shown in Figure~\ref{fig:freeze-tv-sp}.
Stratifying by experience, expert ($n=3$) and novice ($n=14$) operators achieved nearly identical mean quality (4.67 vs. 4.61/5); experts froze at a slightly lower mean percentile (40.1\% vs. 46.8\%), but the difference was not significant ($t(15)=0.457, p=0.654$), likely reflecting the small expert sample. Proximity metrics correlated only weakly and non-significantly with the obstetrician's quality scores: freeze percentile (Spearman $\rho=-0.247, p=0.340$), translational distance ($\rho=-0.291, p=0.257$), and rotational proximity ($\rho=+0.298, p=0.254$; $n=17$), with negligible differences between high-quality ($\geq 4.5/5$) and low-quality ($< 4.5/5$) groups. 

\subsection{Real-Time Performance and Deployment on NVIDIA Clara AGX}
\label{results:clara}

We benchmark the pipeline on a development workstation (NVIDIA A100-SXM4-40GB GPU) and an embedded clinical edge device (NVIDIA Clara AGX). On the workstation, inference took 37.11\,ms for preprocessing (frame extraction, classification, and segmentation) and 5.50\,ms for pose regression, for a total of 42.61\,ms per frame, approximately 23\,Hz. We deploy the inference pipeline on an NVIDIA Clara AGX to assess feasibility for deployment in clinical \ac{us} systems. With both models converted to TensorRT format and no architectural changes, preprocessing/segmentation dropped to 16.08\,ms and pose regression to 9.55\,ms, for a total per-frame inference time of 25.64\,ms, approximately 39\,Hz. 

\section{Discussion}
\label{sec:discussion}

By generalizing to arbitrary, unlabeled non-\acp{sp} (Table~\ref{tab:iou-res}), our semi-supervised approach avoids the impracticality of exhaustive annotation. While supervised baselines (\emph{S-Seg}, AutoFB~\cite{AutoFB}) overfit to \acp{sp} and collapse on arbitrary slices, consistency regularization (\emph{SS-Seg}) shifts the encoder toward shape-invariant features. Reintroducing the classification branch (\emph{Class+SS-Seg}) effectively balances this trade-off, recovering \ac{sp} accuracy while preserving robust non-\ac{sp} generalization. Furthermore, segmentation masking reduces the median translation and rotation errors of the pose regression by 16\% and 15\%, respectively (Table~\ref{tab:posereg-loocv-res}), indicating that focusing the network on the brain suppresses interference from irrelevant anatomy and \ac{us} beam-shape artifacts and supporting the segmentation-first design.

Direct quantitative comparison with prior plane localization methods~\cite{Yeung2021b,Yeung2022} is difficult due to differences in training/test data (synthetic slices vs. real freehand video), evaluation protocols (offline slice accuracy vs. online proximity), and code/data availability. Many also require \ac{3d} volume segmentation masks, a significantly larger labeling burden. Where they share components with our pipeline (e.g., Yeung~\emph{et~al.}~\cite{Yeung2021b} also segment as preprocessing), they could be integrated without altering our core contributions of semi-supervised segmentation and proximity estimation.

While our labeled \ac{sp} training data spans 42 fetuses, the unlabeled non-\ac{sp} images are extracted from only six volumes. Expanding this volumetric dataset is expected to improve the pose regression model's generalization to off-plane variability. 
Extending the pipeline to other \acp{sp} and \acp{ga} would likely degrade the guidance signal: posterior-fossa shadowing near the \ac{tc} \ac{sp} would fragment the \emph{Class+SS-Seg} masks and destabilize pose regression, the lower contrast of the thalami would weaken the signal for the \ac{tt} \ac{sp}, and third-trimester cranial ossification would yield incomplete brain masks and unreliable pose estimates; the second-trimester \ac{tv} plane therefore remains the most reliable target for an initial clinical deployment.

The classification branch serves two complementary purposes. First, it filters the many non-brain frames in clinical video (maternal tissue, probe artifacts, start/end frames) that would otherwise produce spurious proximity measurements. Second, as an auxiliary objective, it pushes the encoder toward brain-discriminative features, incidentally improving segmentation quality (Table~\ref{tab:iou-res}).

Our retrospective analysis suggests that the proximity signal is a step toward a meaningful relative guidance cue. Although no guidance was shown to the sonographers during the recorded scans, operators tend to freeze near, rather than exactly at, the local minima of the proximity trajectory, so the signal aligns with the operator's visual recognition of anatomy even when the manual freeze is physiologically delayed. However, freeze timing remains multifactorial (e.g., visual search, task complexity, operator attention), and geometric proximity alone does not strongly predict the obstetrician's quality scores, which depend on nuanced factors such as acoustic shadows or zoom. A kinematic analysis of these same recordings found that experts move the probe more slowly ($12.9 \pm 3.4$ vs. $24.6 \pm 5.7$ mm/s) and more smoothly than fellows~\cite{lelous2024probe}, suggesting that they prioritize fine-tuning and anatomy verification over rapid navigation, which may explain why a purely geometric metric does not separate the two groups. Thus, while continuous proximity is a geometric prerequisite for reaching the \ac{sp}, final fine-tuning still requires assessing visible landmarks. Existing classification-based methods are similarly incapable of filtering sub-optimal \acp{sp} based on these subtle quality metrics, highlighting an area for future work.

A discrepancy exists between slice-based \ac{loocv} metrics and video-based guidance behavior: horizontal flip augmentation degrades pose accuracy on synthetic slices (Table~\ref{tab:posereg-loocv-res}) due to overfitting to orientation. Because our volumes share a fixed canonical reference frame, a model trained without flip augmentation overfits to this layout, producing artificially low \ac{loocv} errors. By contrast, freehand videos introduce unpredictable probe mirroring and varying scanning directions. Flip augmentation provides the spatial invariance essential for clinical guidance, justifying the degraded slice-level precision in pose regression. Although our 17-scan sample limits the strength of our statistical conclusions, the consistent window-based improvement in proximity ranking motivates further investigation of proximity-based guidance in larger cohorts.

The primary clinical value of our proximity metrics lies in providing actionable guidance for probe manipulation, reinforcing the decoupling of translation and rotation. Translational distance ($d_{trans}$) offers a smooth, actionable trajectory for sliding the transducer toward the correct anatomical area, whereas rapid wrist adjustments cause rotational distance to fluctuate, making it better suited as a background constraint rather than a guidance cue. Fusing these units would artificially inflate the signal noise. At present, the proximity signal is \emph{passive}: the distance is displayed as a numerical value or a continuous 1D plot. This ``distance-only'' presentation avoids the cognitive load of interpreting raw \ac{6d} coordinates and indicates whether the current motion is converging toward the target, but it splits attention between the plot and the \ac{us} feed and does not indicate which direction to move. It is therefore an intermediate step: the next step is active guidance (directional arrows, trajectory traces, or haptic cues), whose usability, together with that of decomposed translation/rotation feedback, warrants dedicated user-centered studies.
 
Real-time operation is critical to maintaining hand-eye coordination. Our optimized Clara AGX deployment (39\,Hz) exceeds the 25--30\,Hz frame rate of commercial \ac{us} scanners, leaving a margin for keeping navigation feedback synchronized with probe motion. The system is thus computationally capable of live execution, but its validation to date is retrospective: a prospective study in which the sonographer sees and reacts to the proximity feedback during a live scan, initially on physical fetal phantoms and then in vivo, remains necessary to establish clinical efficacy.

\section{Conclusion}
\label{sec:conclusions}
This paper presents a sensorless, real-time proximity estimation system for fetal brain \ac{us}, advancing plane localization from binary classification (\ac{sp} detected/not detected) to continuous guidance (distance to the target plane). By combining semi-supervised brain segmentation with \ac{6d} pose regression, our approach is a step toward active, continuous guidance and has the potential to enable earlier and more precise probe adjustments during scanning, linking algorithmic estimation to clinical usability. We foresee initial deployment as an educational tool in phantom-based simulation, where novices receive objective, real-time feedback on their probe trajectory. Clinical translation then requires prospective user studies, first on phantoms and then in live examinations, evaluating acquisition speed, accuracy, and user confidence; multicenter validation across equipment, \acp{ga}, and pathological cases; and extension to the \ac{tt} and \ac{tc} planes together with temporal smoothing and workflow integration.

%
\bibliographystyle{IEEEtran}
\bibliography{tmi}

\end{document}